\documentclass{article}
\usepackage{amsmath,graphicx,mlspconf}

\usepackage{microtype}
\usepackage{graphicx}
\usepackage{subfigure}
\usepackage{booktabs} % for professional tables

\usepackage{bm}
\usepackage{xcolor}
\usepackage[page]{appendix}
\usepackage{enumerate}

\usepackage{lscape}
\usepackage{booktabs}
\usepackage{rotating}
\usepackage{multirow}
\usepackage{longtable}
\usepackage{caption}
\usepackage{float}
\usepackage{tabularx}
\usepackage{ragged2e}
\newcolumntype{Y}{>{\RaggedRight\arraybackslash}X}
\usepackage{pdflscape}
\usepackage{afterpage}

\usepackage{amsmath}
\usepackage{amssymb}
\usepackage{amsthm}
\usepackage{mathtools}
\usepackage{dsfont}

\newtheoremstyle{thmstyle}%                % Name
  {}%                                     % Space above
  {}%                                     % Space below
  {\itshape}%                                     % Body font
  {}%                                     % Indent amount
  {\bfseries}%                            % Theorem head font
  { }%                                    % Punctuation after theorem head
  { }%                                    % Space after theorem head, ' ', or \newline
  {\thmname{#1}\thmnumber{ #2}\thmnote{ (#3)}}%                                     % Theorem head spec (can be left empty, meaning `normal')

\usepackage{textcomp}
\usepackage{sourcecodepro}
\usepackage{listings}
\definecolor{commentgrey}{gray}{0.45}
\definecolor{backgray}{gray}{0.96}
\usepackage[ruled,vlined]{algorithm2e}

\SetCommentSty{algcommstyle}
\newtheoremstyle{algodesc}{}{}{}{}{\bfseries}{.}{ }{}%
\theoremstyle{algodesc}

\DeclareMathOperator*{\argmax}{arg\,max}
\DeclareMathOperator*{\argmin}{arg\,min}

\DeclarePairedDelimiter{\norm}{\lVert}{ \rVert}

\newcommand*{\E}{\mathbb E}
\newcommand*{\R}{\mathbb R}

\newcommand*{\Fcal}{\mathcal F}

\newcommand*{\cond}{\;\ifnum\currentgrouptype=16 \middle\fi|\;}
\newcommand*{\ttilde}{{\raise.17ex\hbox{$\scriptstyle\sim$}}}

\makeatletter
\newsavebox{\mybox}\newsavebox{\mysim}
\newcommand*{\distas}[1]{%
  \savebox{\mybox}{\hbox{\kern3pt$\scriptstyle#1$\kern3pt}}%
  \savebox{\mysim}{\hbox{$\sim$}}%
  \mathbin{\overset{#1}{\kern\z@\resizebox{\wd\mybox}{\ht\mysim}{$\sim$}}}%
}
\makeatother

\makeatletter
\def\moverlay{\mathpalette\mov@rlay}
\def\mov@rlay#1#2{\leavevmode\vtop{%
   \baselineskip\z@skip \lineskiplimit-\maxdimen
   \ialign{\hfil$\m@th#1##$\hfil\cr#2\crcr}}}
\newcommand*{\charfusion}[3][\mathord]{
  #1{\ifx#1\mathop\vphantom{#2}\fi\mathpalette\mov@rlay{#2\cr#3}}
  \ifx#1\mathop\expandafter\displaylimits\fi}
\makeatother

\theoremstyle{thmstyle}
\newtheorem{theorem}{Theorem}[section]
\newtheorem{theorem*}{Theorem}

\newtheorem{corollary*}[theorem*]{Corollary}

\newtheorem{proposition*}[theorem*]{Proposition}

\newtheorem{lemma*}[theorem*]{Lemma}

\theoremstyle{definition}

\newtheorem{definition*}{Definition}
\newtheorem*{definition**}{Definition}

\theoremstyle{plain}

\newtheoremstyle{algodesc}{}{}{}{}{\bfseries}{.}{ }{}%
\theoremstyle{algodesc}
\newcommand*{\mat}[1]{\bm{#1}}
\newcommand*{\zetahat}{\bm{\hat{\zeta}}}
\title{High Dimensional Stochastic Linear Contextual Bandit with Missing Covariates}
\name{%
    Byoungwook Jang$^{\star }$%
    \quad Julia Nepper$^{\dagger}$%
    \quad Marc Chevrette$^{\dagger, \ddagger}$%
    \quad Jo Handelsman$^{\dagger, \ddagger}$%
    \quad Alfred O. Hero III$^{\star \mathsection}$\thanks{This work was partially supported by grants 
ARO W911NF-19-1026 and DE-NA0003921.}%
}
\address{%
    $^{\star}$ Department of Statistics, University of Michigan \quad
    $^{\dagger}$ Wisconsin Institute for Discovery \\
    $^{\ddagger}$ Department of Plant Pathology, University of Wisconsin-Madison \\
    $^{\mathsection}$ Department of EECS, University of Michigan
}

\begin{document}
%\ninept

\maketitle

\begin{abstract}
    Recent works in bandit problems adopted lasso convergence theory in the sequential decision-making setting. 
    Even with fully observed contexts, there are technical challenges that hinder the application of existing lasso convergence theory: 1) proving the restricted eigenvalue condition under conditionally sub-Gaussian noise and 2) accounting for the dependence between the context variables and the chosen actions. 
    This paper studies the effect of missing covariates on regret for stochastic linear bandit algorithms. 
    Our work provides a high-probability upper bound on the regret incurred by the proposed algorithm in terms of covariate sampling probabilities, showing that the regret degrades due to missingness by at most $\zeta_{min}^2$, where $\zeta_{min}$ is the minimum probability of observing covariates in the context vector. We illustrate our algorithm for the practical application of experimental design for collecting gene expression data by a sequential selection of class discriminating DNA probes.
\end{abstract}

\section{Introduction}
High-dimensional linear stochastic bandits have become of increasing interest in 
%many applications including 
recommendation systems, healthcare, and experimental design.
Particular attention has been on linear bandits where only a small subset of the covariates is correlated with rewards. In a linear bandit, the learner observes a context variable $X_t \in \mathbb{R}^{K\times d}$ at round $t$, where each arm $i$ is associated with a given feature vector $X_{t,i} \in \mathbb{R}^d$, the $i$-th row of $X_t$. 
Then, based on the chosen arm $a_t$ at time $t$, the learner observes a noisy reward $\hat{r}_t = X_{t,a_t} \beta^* + \varepsilon_t$ for a fixed and unknown parameter vector $\beta^* \in \R^d$. The objective is to determine a policy for choosing $a_t$ to maximize the expected reward based on past observations $\{X_{i, a_i}, \hat{r}_i\}_{i=1}^{t-1}$. 

Frequently, an additional sparsity assumption is imposed on $\beta^*$.
Sparsity is an effective constraint when the feature space is high-dimensional, but only a subset of features are correlated with the expected reward.
In this case, it is natural to adopt the lasso regression framework where the reward parameter $\beta^*$ satisfies $s_0 = \norm{\beta^*}_0 \ll d$. This results in variants on the lasso bandit including the sparse agnostic (SA) bandit \cite{oh2021sparsity}, which is then extended to the sparse agnostic with missingness (SAM) bandit in this paper. 

Missing values are common in contextual bandits \cite{tewari2017ads}, motivating this work from a practical standpoint.
For example, in a clinical application of warfarin dosing with patient data \cite{international2009estimation}, missing values for certain genotypes are prevalent in the context vectors.
These can be imputed based on demographic or other genotype information when available.

This work adopts the regression with missing data framework \cite{lohwainwright12} to unbias the covariance matrix when there are missing values.
% to establish tight convergence for stochastic linear bandits with missing covariates. 
Specifically, we adopt the missing completely at random (MCAR) model.
In this model, variables in the context vector $X_{t,i} \in \R^d$ of the $i$-th arm at time $t$ are observed with probabilities $\bm{\zeta} = [\zeta_1, \dots, \zeta_d] \in [0, 1]^d$.
Mathematically, the observed context with missing entries can be written as $Z = X \odot U$, where $\odot$ is a Hadamard product and $U_{t,i} \in \{0, 1\}^d$ is a $Ber(\zeta_i)$ independent of both $X_{t,i}$ and $\varepsilon_i$. 
% In high-dimensional statistics, a plug-in-estimator was introduced in \cite{lohwainwright12} for linear regression problems with additive and/or multiplicative noise. The authors of \cite{fan2019precision} tackled a similar problem for estimating sparse precision matrices.

The difficulty in establishing convergence of lasso bandits has been highlighted in \cite{oh2021sparsity,bastani2020online, kim2019doubly}. This is due to the fact that the observation noise $\varepsilon_t$ associated with successive pulls of the chosen arms is not i.i.d. 
As first addressed in \cite{bastani2020online}, the sequence $\varepsilon_t X_t$ can be shown to be a Martingale difference sequence under mild conditions. 
This enables the application of tail inequalities to the lasso estimator. 
The authors of \cite{kim2019doubly} used the  Martingale approach to perform regret analysis of the Doubly-Robust (DR) Lasso bandit, and \cite{oh2021sparsity} applied them to the sparse agnostic (SA) Lasso bandit. 

The convergence results \cite{oh2021sparsity} relied on technical conditions, namely the compatibility condition and the restricted eigenvalue condition.
In order to establish these conditions in bandit problems, the oracle lasso convergence theory in \cite{van2009conditions} was modified using a Martingale concentration inequality.
This requires positive-definiteness of the context covariance matrix, a condition that is often violated when there are missing values in the observed contexts.
This paper relaxes this requirement, allowing us to prove convergence when there are missing values in the context variables in stochastic linear bandit problems. 

To the best of our knowledge, this is the first paper to propose a lasso bandit with missing covariates in addition to providing theoretical guarantees. Our work establishes restricted lower- and upper-restricted eigenvalue (RE) conditions on the adjusted sample covariance matrix when the observation noises $\varepsilon_t$ are dependent on the past observations. 
We show that missingness in covariates inflates the regret bounds by a factor inversely proportional to the squared minimum sampling probability $\zeta_{min}^2$, which is inversely proportional to missingness.

\subsection{Related Work}
\textbf{Sparse Linear Contextual Bandit:} 
Interest in sparse linear bandits in high-dimension began with \cite{abbasi2012online}, \cite{carpentier2012bandit}, and continued with \cite{gilton2017sparse} and \cite{bastani2020online}. 
Recently, \cite{kim2019doubly} applied sparse linear structure to the stochastic linear bandit problem and incorporated a doubly-robust approach to prove convergence. 
The regret analysis in \cite{kim2019doubly} mainly adopts the procedure in \cite{bastani2020online} that extended the standard lasso convergence results to online regression with non-i.i.d. samples. 
The authors of \cite{oh2021sparsity} introduced sparse agnostic Lasso bandit under the balanced covariance assumption to circumvent the dependency problem.
We relax the positive semi-definite assumption on the covariance. 

\textbf{Missing data in regression and covariance estimation:} 
Traditional methods introduced in \cite{stadler2012missing} worked with the EM algorithm to perform statistical inference for missing data. 
However, even in the batch setting where they do converge, EM algorithms often converge slowly. 
M-estimators developed in \cite{lohwainwright12} were designed in mind to cope with missing and corrupted data by making a simple adjustment to the sample covariance matrix. 
The adjusted sample covariance estimates, however, are not necessarily positive semi-definite, which makes the resulting likelihood non-convex. 
In \cite{lohwainwright12}, it was proved that the local and global optima of the non-convex lasso problem have comparable mean squared error.
Thus, a simple projected gradient algorithm for the non-convex lasso objective under missing data is sufficient to guarantee the convergence of these estimators. 
We apply the lasso framework of \cite{lohwainwright12} to obtain an integrated and principled solution to the sparse agnostic with missingness (SAM) bandit. 

% \textbf{Experimental Design with Bandit}:

\section{Problem Setup}
We introduce our formulation of the linear contextual bandit problem under missing data and motivate the proposed estimator. 

\textbf{Missing covariates:} In a sparse stochastic linear bandit, the reward for pulling the $i$-th arm at time $t$ is of the form $r_{t,i} = X_{t,i} \beta^*$, for $i \in [K]$, given the covariates $X_{t,i} \in \R^d$, called context variables. 
Over the time steps $[T]$, the learner estimates the unknown regression parameter $\beta^* \in \R^d$, which is assumed to be sparse.
%, reflecting the lack of correlation between the reward and a subset of the covariates. 
In a typical stochastic linear bandit problem, after pulling arm $a_t$, we observe a reward $\hat{r}_{t}$, linked with context vector $X_{t,a_t} \in \R^d$, via the noisy linear model 
\begin{equation}
\label{eq:noisy_linear_reward}
    \hat r_{t} = X_{t,a_t} \beta^* + \varepsilon_{t} \qquad t \in [T], a_t \in [K],
\end{equation}
where $\varepsilon_{t} \in \R$ is the observation noise independent of $X_{t,i}$. 
Instead of directly observing $X_{t,i}$, we observe $Z_{t,i} = [Z_{t, i1}, \dots, Z_{t, id}]$ with missing entries defined as follows
\begin{equation}
    \label{eq:missing_observation}
    Z_{t,ij} = 
    \begin{cases} 
        X_{t,ij} & \text{if the entry is not missing} \\
        0 & \text{if the entry is missing}
    \end{cases}.
\end{equation}
Equivalently, the learner observes the context for the $i$-th arm at time $t$ as $Z_{t,i} = X_{t,i} \odot U_{t,i}$, where $U_{t,i} \in \{0, 1\}^d$ 
%is  vector of i.i.d. Bernoulli random variables, independent of $X_{t,i}$, 
and $\odot$ is the Hadamard product. 
Each entry $U_{t,ij}$ is an independent Bernoulli random variable with sampling probability parameter $\zeta_j$ for the $j$-th covariate of the context vector $X_{t, i}$
The estimation goal is to recover $\beta^*$ as we sequentially receive the context vectors $Z_{t,i}$ with missing entries. 

\textbf{Sparse Agnostic with Missingness bandit:} 
%\subsection{Multi-armed Bandit with Missing Covariates}
The learner must decide which of $K$ available arms to pull based on the observed contexts $Z_{t,i} = X_{t,i} \odot U_{t,i} \in \R^d$, $i \in [K]$. 
That is, at time $t$, we observe $\mat{Z}_t \in \R^{K \times d}$ matrix-variate data and the missingness pattern $\mat{U}_t \in \R^{K \times d}$, where the $i$-th row of these variables corresponds to the $i$-th arm.
Based on the context variables, the learner pulls an arm and incurs a reward.
% Defining the matrix $X_t \in \R^{K \times d}$ of all contexts at time $t$, the DR Lasso bandit of \cite{kim2019doubly} addresses the missingness of rows of $X_t$, due to the selection of one arm per time $t$, while this paper addresses missingness in both rows and columns. 
Note that when $\zeta_j = 1$ for all covariates $j \in [d]$, our problem setting is the same as the fully observed case of \cite{kim2019doubly} and \cite{oh2021sparsity}. 

The policy we adopt in this paper is to pull the arm maximizing the estimated reward
\begin{equation}
\label{eq:plug-in-estimation-policy}
    a_t = \argmax_{i\in [K]} (Z_{t,i} \oslash \zetahat) \hat\beta_{t-1}, \qquad t \in [T],
\end{equation}
where $\oslash$ is element-wise division, $\zetahat \in \mathbb{R}^d$ is an empirical estimate of the sampling probabilities based on the observed $U_t$'s, and $\hat{\beta}_t$ is an estimate of $\beta^*$. 
The policy defined by \eqref{eq:plug-in-estimation-policy} is called the plug-in-estimation policy \cite{lohwainwright12}. 

We define the optimal arm at time $t$ as 
\begin{equation}
\label{eq:oracle_action}
    a^*_t = \argmax_{1\leq i \leq K} X_{t,i} \beta^*, \qquad t \in [T],
\end{equation} 
and the \textit{regret(t)} as the difference between the expected reward of the optimal arm and the expected reward of the chosen arm at time $t$ based on \eqref{eq:plug-in-estimation-policy}.
\begin{equation*}
    \begin{aligned}
        regret(t) 
        & = \E[r_{t, a^*_t} - r_{t, a_t} | \{X_{i,t}\}_{i=1}^K, a_t, U_t] \\
        & = X_{t,a^*_t}\beta - X_{t, a_t}\beta, \qquad t \in [T],
    \end{aligned}
\end{equation*}
where $r_{t, a_t^*}$ and $r_{t, a_t}$ are the maxima achieved in \eqref{eq:plug-in-estimation-policy} and \eqref{eq:oracle_action}.
The learner aims to minimize the cumulative regret over $T$ steps. 

%\textbf{Notation:} 
For the rest of the paper, we define the filtration $\Fcal_{t-1}$ as the union of all observations up to time $t-1$ including rewards, missingness patterns, and contexts:
\begin{equation*}
\begin{aligned}
    \Fcal_{t-1} = \{ (Z_\tau, \hat{r}_{\tau, a_\tau}, U_\tau) \}_{\tau=1}^{t-1}.
\end{aligned}
\end{equation*}
Given $\Fcal_{t-1}$, the learner selects the arm $a_t$ according to the current estimate $\hat{\beta}_t$.
%the probability $\pi_t = [\pi_{t, 1}, \dots, \pi_{t, K}]$, where $\pi_{t, i} = \mathbb{P}[a_t = i | \Fcal_{t-1}]$. 

\section{SA Lasso bandit with missing covariates}
We introduce covariate missingness into the SA Lasso bandit \cite{oh2021sparsity} when the contexts are corrupted with missing values. 
We analyze the regret when the plug-in-estimation policy \eqref{eq:plug-in-estimation-policy} is used in SA with missingness (SAM) bandit algorithm. 
The same plug-in adjustment procedure and regret analysis can be applied to the lasso bandit and to the DR bandit with missingness (DRM) but we omit the details here. 

\subsection{Bandit with missing covariates}
\label{section:sa_lasso_bandit}
Compared to \cite{kim2019doubly}, where the uneven sampling of the covariates is addressed by taking the average of the contexts at each round and calculating the corresponding pseudo-rewards, \cite{oh2021sparsity} introduced the SA Lasso bandit by making an additional assumption on the distribution of the contexts such that the covariance matrix $\Sigma_t = \E [X_t^T X_t | \mathcal{F}_{t-1}]$ behaves sufficiently well to converge to the marginal context covariance matrix $\Sigma = \E [X_t^T X_t]$. 
More specifically, \cite{oh2021sparsity} showed that
\begin{equation*}
    \sum_{i=1}^k \E_{\mathcal{X}_t} \left [ X_{t,i}^T X_{t,i} 
    \mathbf{1}(X_{t,i} = \argmax_{X\in \mathcal{X}_t} X \beta^*) \right] 
    \succcurlyeq (2 \nu C_{\mathcal{X}})^{-1} \Sigma,
\end{equation*}
under the balanced covariance assumption (Assumption 5) stated later in this paper, where $\bm{X}_t = [X_{t, 1}, \dots, X_{t, K}]^T$ (See Lemma 3 and Lemma 10 of \cite{oh2021sparsity} for more details). 
Here we will show that under the balanced covariance assumption, we can achieve a similar result to \cite{oh2021sparsity} even in the presence of missing covariates.

For the SA Lasso bandit with missing covariates algorithm, we directly use the observed rewards $\hat{r}_{t}$ and the incompletely observed contexts $Z_{t, a_t} = X_{t, a_t} \odot U_{t, a_t}$.
Thus, we define $\mat{Z}_t = [Z_{1, a_1}, \dots Z_{t, a_t}]\in \mathbb{R}^{t \times d}$ and $\bm{\hat{r}}_t = [\hat r_{1}, \dots, \hat r_{t}] \in \mathbb{R}^t$, where $Z_{\tau, a_\tau} = X_{\tau, a_\tau} \odot U_{\tau, a_\tau}$ and $\hat r_\tau = X_{\tau, a_\tau} \beta + \varepsilon_\tau$.

\subsection{Lasso estimation with adjusted covariance matrix}
Based on the observed $\mat{Z}_t$, we optimize
\begin{equation}
    \label{eq:lasso_problem}
    \hat \beta_{t} \in \argmin_{\norm{\beta}_1 < R} \left \{ \frac{1}{2} \beta^T \hat{\Gamma}_{miss, t} \beta - \langle \hat{\gamma}_{miss, t} , \beta \rangle + \eta_t \norm{\beta}_1\right \}
\end{equation}
where $\norm{\cdot}_1$ is an $\ell_1$ norm and 
\begin{equation}
    \label{eq:Gamma_definition}
    \begin{aligned}
    & \hat{\Gamma}_{miss, t} = \left (\frac{1}{t} \mat{Z}_t^T \mat{Z}_t \right) \oslash \hat{M} \\
    & \hat{\gamma}_{miss, t} = \left ( \frac{1}{t} \mat{Z}_t^T \mathbf{r}_t \right) \oslash \zetahat \\
    & \hat{M}_{ij} = 
    \begin{cases}
        \zetahat_i & \text{if } i = j \\
        \zetahat_i \zetahat_j & \text{if } i \neq j
    \end{cases}
    \end{aligned},
\end{equation}
where $\zetahat \in \R^d$ is the sampling probability of each of the covariates. 

Due to the missing value, $\hat{\Gamma}_{miss, t}$ contains negative eigenvalues that make \eqref{eq:lasso_problem} non-convex and unbounded from below. In order to circumvent this problem, \cite{lohwainwright12} introduced the additional $\ell_1$ constraint on $\beta$ for noisy and missing data. Our analysis adopts this estimator for the high-dimensional stochastic linear bandit problem. \cite{loh2015regularized} further extended this problem for non-convex regularization in the batch setting. 

While applying the approach of \cite{lohwainwright12} to the bandit problem seems simple enough, several challenges arise due to the sequential-decision making setting of the linear bandit. 
As the noise $\varepsilon_{t}$ is not i.i.d., we cannot directly apply the convergence results from \cite{lohwainwright12} to the lasso bandit with covariate missingness.
Nonetheless, in Theorem~\ref{theorem:sa_lasso_regret} below, we can establish convergence. 

% We show that the optimal regularization path depends on the minimum sampling probability of the covariates $\zeta_{min} = \min_j \zetahat_j$. Our theorem shows that $\eta_t$ should scale with $\frac{\log d}{t \cdot \zeta_{min}^2}$. 
% This regularization path agrees with the noiseless setting in \cite{oh2021sparsity}. Intuitively, the effective sample size of the covariates is $t \cdot \zeta_{min}^2$, which is the number of time steps needed to reliably estimate the off-diagonal entries of $\E[X_t^T X_t | \Fcal_{t-1}]$.
% It will be easily seen that all of our results will reduce to \cite{oh2021sparsity} for the case that $\zeta_j = 1$ for all $j \in [K]$. 

\section{Algorithm}
Algorithm~\ref{alg:sa_lasso_missing} solves the lasso bandit problem under the covariate missingness.
The key differences compared to the fully-observed counterpart are 1) the use of adjusted plug-in estimators $\hat \Gamma_{miss, t}$ and $\hat \gamma_{miss, t}$ and 2) the theoretically justified regularization parameter $\eta_t$.
An additional tuning parameter $R$ is introduced due to the non-convexity of the problem \eqref{eq:lasso_problem} in order to constrain $\beta$ to be in an $\ell_1$ ball. This is motivated by a similar condition proposed in \cite{lohwainwright12} and \cite{rudelson2017errors}.

% \onecolumn
\begin{algorithm}[ht]
\caption{SA with missingness (SAM) bandit} 
\label{alg:sa_lasso_missing}
  \SetAlgoLined
  \KwIn{$\eta_1$, $R$}
%   \KwOut{$\hat{\Theta}$}
  Initialize $\beta_0 = 0$, $\zetahat_0 = 1$
  
  \For{$t=1, \dots, T$}{
    Observe contexts $Z_t \sim \mathcal{P}_{K\times d}$  \\
    and the missing pattern $U_t$ 
    
    Update $\zetahat_t = \zetahat_{t-1} + \frac{1}{t} \left(\frac{1}{K}\sum_{i=1}^K U_{t,i} - \zetahat_{t-1}\right)$
    
    Pull arm $a_t = \argmax_{i \in [k]} (Z_{t,i} \oslash \hat \zeta_t ) \hat \beta_t$
    
    Observe $\hat{r}_{t}$ for the arm $a_t$
    
    Update $\eta_t = \eta_1 \sqrt{\frac{4 \log (t\; \zeta_{min}^2) + \log d}{t \; \zeta_{min}^2}}$
    
    % $\hat{\Gamma}_{miss, t} = \left (\hat{\Gamma}_{miss, t-1} + \frac{1}{t} \left(Z_{t, a_t} Z_{t, a_t}^T -\hat{\Gamma}_{miss, t-1}  \right) \right ) \oslash M_t$
    Updated $\hat{\Gamma}_{miss, t}$ and $\hat{\gamma}_{miss,t}$ based on \eqref{eq:Gamma_definition}
    
    Update $\hat \beta_t$ based on \eqref{eq:lasso_problem}
  }
\end{algorithm}
For the plug-in estimator in Algorithm~\ref{alg:sa_lasso_missing}, we consider the constrained program as introduced in \cite{loh2015regularized}, given $\hat{\Gamma}_{miss, t}$ and $\hat{\gamma}_{miss, t}$,
\begin{equation}
    \label{eq:constrained_regression_missing_l1}
    \begin{aligned}
    \hat{\beta}_t \in \argmin_{\norm{\beta}_1 \leq R } 
    & \Big \{ \frac{1}{2} \beta^T \widehat{\Gamma}_{miss, t} \beta  - \langle \widehat{\gamma}_{miss, t}, \beta \rangle + \eta_t \norm{\beta}_1 \Big \}
    \end{aligned},
\end{equation}
for some constant $R = b \sqrt{s_0}$, where $b = \max_{i\in [d]} \beta^*$. 
As we do not have a priori knowledge of the sparsity level $s_0$, $R$ is a user-defined parameter.
While typical gradient descent methods fail to converge to a global optimum due to local minima, \cite{lohwainwright12} proved that a simple projected gradient descent algorithm for \eqref{eq:constrained_regression_missing_l1} converges within the statistical tolerance of the global optimum even when $\widehat{\Gamma}_{miss, t} \not \succcurlyeq 0$. 

% To iteratively solve \eqref{eq:constrained_regression_missing_l1}, we apply the following Lagrangian update with $\ell_1$-ball penalty parameter $\mu$ and sparsity parameter $\eta_t$.
% \begin{equation}
%     \label{eq:projected_gradient_constrained_l1}
%     \begin{aligned}
%     \beta^{i+1} = 
%     & \argmin_{\norm{\beta}_1 \leq R} \Big \{ \mathcal{L}(\beta^i) + \langle \nabla \mathcal{L}(\beta^i), \beta - \beta^i\rangle \\
%     & \qquad \qquad \qquad + \frac{\mu}{2} \norm{\beta - \beta^i}_2^2 
%     + \lambda_t \norm{\beta}_1 \Big \}
%     \end{aligned},
% \end{equation}
% where $\mu>0, \eta_t > 0$, $\mathcal{L}(\beta) = \frac{1}{2} \beta^T \widehat{\Gamma}_{miss, t} \beta - \langle \widehat{\gamma}_{miss, t}, \beta \rangle$ is the likelihood, and $\nabla \mathcal{L}(\beta) = \widehat{\Gamma}_{miss, t} \beta - \widehat{\gamma}_{miss, t}$ is its gradient. 

\section{Regret analysis under missing data}
Here we give an upper bound on the regret of the proposed SAM bandit defined by Algorithm~\ref{alg:sa_lasso_missing}. This analysis equally applies to the DRM bandit.
% We emphasize that the regret analysis is not limited to our problem but can be derived for any linear bandits using Proposition~\ref{thm:lasso_missing_martingale_stat_error}. 
% This key result extends Proposition 1 in \cite{bastani2020online} to the case where there are missing values. 
The distributional assumptions on the covariates and the reward noise include the bounded feature set and reward function parameter (\textbf{A1}), i.i.d. on the context variables $X_t \in \mathbb{R}^{K\times d}$ (\textbf{A2}), and the sub-gaussianity of the error $\varepsilon_t = \hat{r}_{t, a_t} - X_{t, a_t} \beta^*$ (\textbf{A3}). A1-3 are standard assumptions in the stochastic linear bandit literature \cite{bastani2020online, kim2019doubly,oh2021sparsity} to derive the upper bound for the cumulative regret.
% \begin{assumption}[Feature set and parameter]
% \label{assumption:feature}
%     There exists a positive constant $x_{max}$ such that $\norm{X_{t,i}}_2 \leq x_{max}$ for all $X_{t,i} \in \R^d$ and a positive constant $b$ such that $\norm{\beta^*}_2 \leq b$ and $\norm{\beta^*}_0 = s_0$.
% \end{assumption}

% \begin{assumption}[i.i.d. context] 
% \label{assumption:iid_context}
%     The context variables $X_t \in \R^{K \times d}$ are i.i.d. and follow matrix-variate distribution $\mathcal{P}_X$ at every time $t$:
% \end{assumption}

% \begin{assumption}[Sub-Gaussian error]
% \label{assumption:sub_gaussian_error}
%     The error $\varepsilon_{t} = \hat{r}_{t,a_t} - X_{t,a_t} \beta^*$ is $\sigma_\varepsilon$-sub-Gaussian adapted to $\Fcal_{t-1}$ for some $\sigma_\varepsilon > 0$. 
%     In other words, for every $\alpha \in \R$, $\mathbb{E}[e^{\alpha \varepsilon_{t}}|\Fcal_{t-1}] \leq e^{\sigma_\varepsilon^2 \alpha^2/2}$. 
% \end{assumption}
%\begin{assumption}[Sub-Gaussian error]
%\label{assumption:sub_gaussian_error}
    %The error $\varepsilon_{t} = \hat{r}_{t,a_t} - X_{t,a_t} \beta^*$ is $\sigma_\varepsilon$-sub-Gaussian adapted to $\Fcal_{t-1}$ for some $\sigma_\varepsilon > 0$. 
    %In other words, for every $\alpha \in \R$, $\mathbb{E}[e^{\alpha \varepsilon_{t,i}}|\Fcal_{t-1}] \leq e^{\sigma_\varepsilon^2 \alpha^2/2}$. 
%\end{assumption}

We additionally involve the compatibility condition on the true Gram matrix $\Sigma := \frac{1}{K} \mathbb{E}[X ^T X]$. This is a standard assumption in high-dimensional regression literature \cite{van2009conditions} and stochastic linear bandit problems \cite{bastani2020online, kim2019doubly, oh2021sparsity,wang2018minimax}. 
Before we define the compatibility condition, we first define the active set $S_0 = \{ j: \beta_j^* \neq 0 \}$ as the set of indices that correspond to non-zero values of $\beta_j^*$. Thus, the true $\beta^*$ can be divided into $\beta_{j, S_0}^* = \beta_j \mathds{1}( j \in S_0) \text{ and } \beta_{j, S_0^c}^* = \beta_j \mathds{1}(j \not \in S_0)$.

Let $\mathbb{C}(S_0)$ be the set of vectors $\beta\in \R^d$ defined as 
\begin{equation}
    \mathbb{C} (S_0) = \{\beta \in \R^d | \norm{\beta_{S_0^c}}_1 < 3 \norm{\beta_{S_0}}_1 \}.
\end{equation}
Then, we can define the compatibility condition.

\noindent\textbf{A3 (Compatibility Condition)}: \textit{For an active set $S_0$, there exists a compatibility constant $\phi^2 > 0$ such that 
$\phi_0^2 \norm{\beta_{S_0}}_1^2 \leq s_0 \beta^T \Sigma \beta \quad \forall \beta \in \mathbb{C}(S_0)$}
% The compatibility condition generalizes the positive-definite assumption on the population covariance matrix and ensures that a Lasso estimator will converge to the true parameter $\beta$ with high probability as the sample size grows to infinity. 
% While the compatibility condition seems technical at first sight, it allows us to bound the $\ell_1$-norm with the $\ell_2$-norm. 
% It can be easily seen that the positive-definite assumption in OLS satisfies the compatibility condition with $\phi_0 = \sqrt{\lambda_{min}(\Sigma)}$.
% Despite the compatibility condition on the true $\Sigma$, our unbiased estimator for $\Sigma$ always contains negative eigenvalues when there are missing covariates in the context vector. Thus, we later introduce restricted strong convexity to prove convergence.

\noindent\textbf{A4 (Relaxed symmetry)}: \textit{For a joint distribution $\mathcal{P}_X$, there exists $\nu < \infty$ such that $\frac{\mathcal{P}_X(- \mathbf{x})}{\mathcal{P}_X(\mathbf{x})} \leq \nu$ for all $\mathbf{x} \in \mathbb{R}^d$.}
The relaxed symmetry assumption is satisfied by a wide range of distributions including the Gaussian distribution and the uniform distribution. Note that for symmetric distributions, A4 is satisfied with $\nu = 1$. 

\noindent\textbf{A5 (Balanced covariance)}: \textit{Consider a permutation $(i_1, \dots, i_K)$ of $(1, \dots, K)$. For any integer $k \in \{2, \dots, K-1\}$ and fixed vector $\beta$, there exists $C_{\mathcal{X}}$ such that}
    \begin{equation*}
    \begin{aligned}
        & \E \left[ X_{i_k}^T X_{i_k} \mathds{1} (X_{i_1} \beta^* < \dots < X_{i_K} \beta^*) \right ] \\
        & \preccurlyeq C_{X} \E \left[ (X_{i_1}^T X_{i_1} + X_{i_K}^T X_{i_K}) \mathds{1} (X_{i_1} \beta^* < \dots < X_{i_K} \beta^*) \right ] \\
    \end{aligned}
    \end{equation*}
In bandit problems, the sample covariance governing the observation at time $t$ is $\mathbb{E}[X^T X| \Fcal_{t-1}]$. As we are selecting arms $a_t = \argmax_{i \in [K]} (Z_{t, i} \oslash \zetahat) \beta_t$ based on the current estimate $\beta_t$, 
the SA bandit algorithm may undersample part of the distribution. 
As introduced in \cite{oh2021sparsity}, the balanced covariance assumption implies that we can control the covariance matrix based on the extreme selections of the arms. 

For example, if the arms are completely correlated, $C_X$ is constant independent of dimensions. In a more general setting, \cite{oh2021sparsity} proved that the balanced covariance condition is satisfied with $C_X = \binom{K-1}{K_0}$ with $K_0 = \lceil \frac{K-1}{2}\rceil$ when the arms are independent and identically distributed from Gaussian distribution. 
The balanced covariance condition was first introduced in \cite{oh2021sparsity}, and the value of $C_X$ gives insight into the behavior of the population covariance matrix.

\begin{theorem}[SA Lasso bandit with missing values]
\label{theorem:sa_lasso_regret}
Suppose Assumption 1, 2, 3, 4 and 5 hold. Then, for some constant $c_0, c_1 > 0$, the cumulative regret of the SA Lasso bandit with missing values is $\mathcal{O} \left(\frac{1}{\zeta_{min}^2} \sqrt{s_0 T\log(dT) }\right )$ with probability at least $1 - c_0 \exp (-c_1 \log d)$.
\end{theorem}
As pointed out in \cite{oh2021sparsity}, the learner does not have to go through the exploration phase to achieve this convergent regret. 
This is due to the balanced covariance assumption (A5). 
Note that, compared to the fully observed setting, the regret is increased by $\frac{1}{\zeta_{min}^2}$ due to the missing covariates. 
This matches our intuition that extra arm pulls are required to accurately estimate the off-diagonal entries of $\widehat{\Gamma}_{miss, t}$ when the covariates are missing.

%%%%%%%%%%%%%%%%%%%%%%%%%
%Need to update the simulations and edits
\section{Simulation study}
\label{section:simulation}
\begin{figure}[h!] \centering
    \begin{subfigure} \centering
        \includegraphics[width=0.9\linewidth]{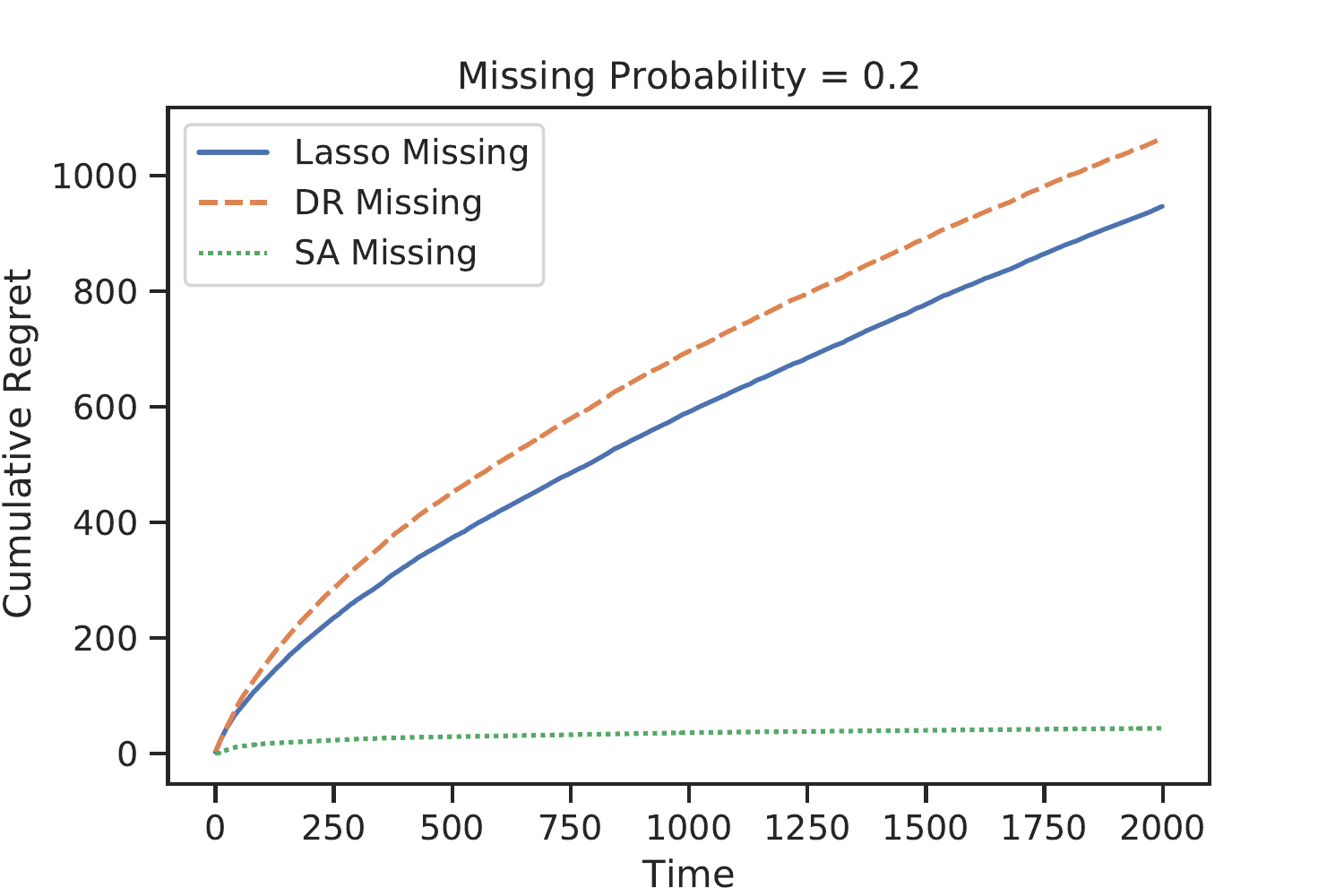}
    \end{subfigure}
    \vspace{-10pt}
    % \begin{subfigure} \centering
    %     \includegraphics[width=0.9\linewidth]{figures/method_comparison_20_200_0.6.pdf}
    % \end{subfigure}
    \caption{Cumulative regret over time for missingness probability $1-\zeta_j = 0.2$ for $j \in [K]$, $k = 20$, and $d=200$ for the proposed modifications of the lasso bandit, the DR Lasso bandit, and the SA Lasso bandit with the missingness adjusted estimator $\hat{\Gamma}_{t, miss}$ and $\hat{\gamma}_{t, miss}$.}
    \label{fig:bandit_missing_comparison}
\end{figure}

\begin{figure}[h!] \centering
    \begin{subfigure} \centering
        \includegraphics[width=0.8\linewidth]{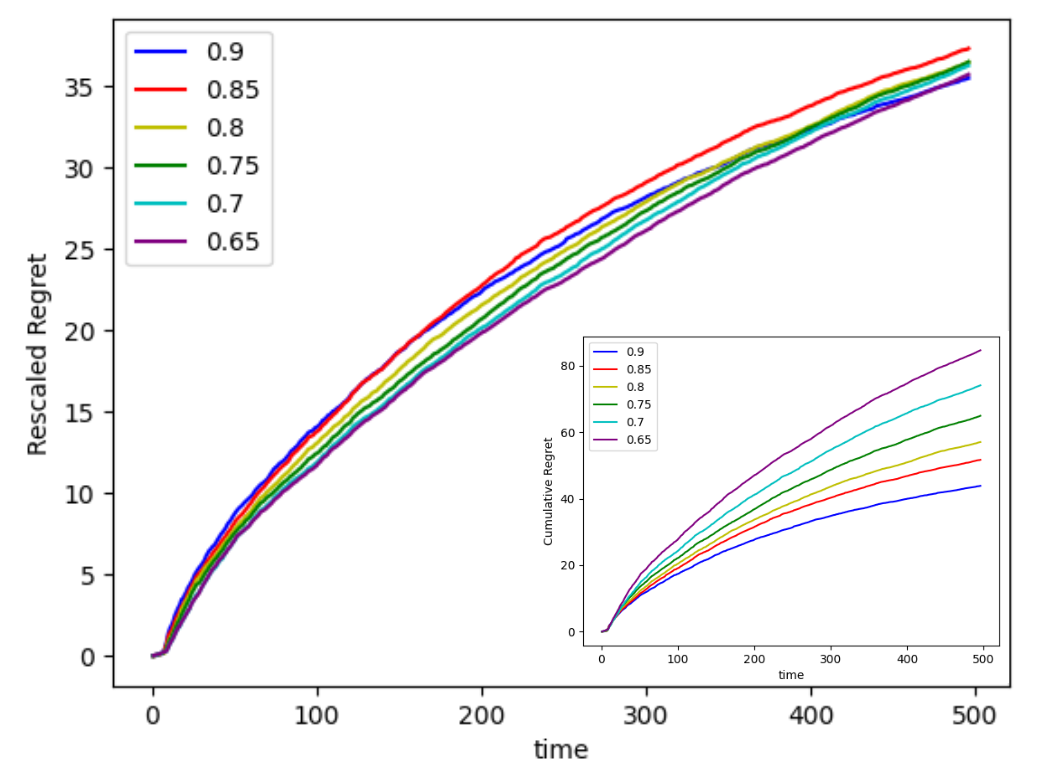}
    \end{subfigure}
    % \begin{subfigure} \centering
    % \vspace{-15pt}
    %     \includegraphics[width=0.95\linewidth]{figures/SA_Lasso_Missing_rescaled.png}
    % \end{subfigure}
    
    \caption{Insert on lower right corner is \textit{regret(t)} to the proposed method over varying $\zeta_j \in [0.65, 0.9]$, $k=20$, and $d=200$. After normalization according to $regret(t) \cdot \frac{\zeta_{min}^2}{\sqrt{s_0 T \log (dT)}}$ validates the rates predicted by Theorem~\ref{theorem:sa_lasso_regret}. }
    \label{fig:sa_bandit_missing_rescaled}
\end{figure}

\textit{\textbf{Simulation}}: We conduct simulations to evaluate the improvement in cumulative regret based on our modifications to the lasso bandit, the DR lasso bandit, and the SA lasso bandit (Figure~\ref{fig:bandit_missing_comparison}). 
% The details of the DR lasso modification are included in Appendix~\ref{appendix:dr_lasso_bandit}.
We set $k=20$, $d=200$ and the missing probability $1-\zeta_j \in \{0.65, 0.9\}$ for all $j \in [K]$.
The sparsity was set to $s_0 = c \sqrt{d}$, where $c$ is constant.  
At each round $t$, we generate the true contexts $X_t \in \R^{K \times d}$, where $X_{t, i} \sim N(0, \Sigma)$ and $\Sigma$ is a Toeplitz matrix. 
The learners observe $Z_t = X_t \odot U_t$, where all entries $U_{t, ij} \sim Ber(\zeta_j)$. 
We could have also varied the sampling probability $\zeta_j$ for each covariate, but Theorem~\ref{theorem:sa_lasso_regret} shows that only the minimum sampling probability plays a role in the convergence rate. 
In this setting, the Gaussian model for $X_t$ satisfies the symmetry condition (A4). 
Lastly, we generate $\varepsilon_{t}$ from the normal distribution, and the reward is observed based on the approximation $Z_t \oslash \zetahat$ of the context $X_t$.
Figure~\ref{fig:sa_bandit_missing_rescaled} shows the regret performance of our algorithm over varying missing probability $1-\zeta_j \in [0.65, 0.9]$.

%\vspace{5pt}
\noindent\textit{\textbf{Case Study: Experimental Design}}:
Understanding the interactions among microbiomes in microbial communities is important to applications in health, ecology, and antibiotic design. 
Recently, \cite{lozano2019introducing} introduced a model microbiome community for the rhizosphere, called THOR or BFK, that combines three microbial species \textit{Bacillus cereus (B), Flavobacterium johnsoniae (F)}, and \textit{Pseudomonas koreensis} to study the complex interactions between them under different conditions (classes). 
We use data from this model to illustrate the application of the proposed contextual bandit with missingness to the sequential design of experiments for discovering the gene probes that best discriminate between two experimental conditions: a BFK community having a wildtype strain of K (class 1) vs a community having a mutant strain of K (class 2). 

\textbf{Gene probe Selection Problem}: 
Often only a few key genes among the thousands of genes play important roles in the behavior of microbial species under varying experimental conditions.
However, simultaneously collecting all available gene expression data for multiple experiments could be costly for researchers. 
We reformulate the DNA probe selection problem as a sequential design problem using contextual bandits. The objective is to sequentially select gene probes (arms) to discover a few genes that best discriminate between the classes. We aim to establish proof of concept that such discriminative genes can be discovered sequentially without the need to sequence the entire genome at once. 

\textbf{Dataset}: Experimental microbiome data was collected and processed in the lab of one of the co-authors.  %by Handelsman lab in the University of Wisconsin. 
We performed gene sequencing on each species, yielding the three gene expression datasets for species $B, F,$ and $K$. 
We applied log-transformation to these expression data and denote the final data as $\mat{X}_i \in \R^{n_i \times (m_1 + m_2)}$ for $i \in \{B, F, K\}$, where $n_i$ is the number of genes for species $i$ and $m_1$ and $m_2$ are the number of replicates (samples) for conditions (classes) 1 and 2, respectively. 
There are $6179$ gene probes for \textit{Bacillus}, $5198$ genes in \textit{Flavobacterium}, and $5864$ genes in \textit{Pseudomonas} with $m_1=38$ and $m_2=34$. 

\textbf{Bandit Formulation}: We formulate the sequential gene selection problem with the contextual bandit having the following components
\vspace{-2pt}
\begin{itemize}
    \item \textit{Arms}: The arms correspond to the genomes of $B, F, K$ for the three species, denoted $\mat{X}\in \mathbb{R}^{k\times(m_1+m_2)}$ where $k$ represents the number of selectable DNA probes, which depends on $B$, $F$, or $K$. 
    \vspace{-5pt}
    \item \textit{Covariates}: The covariates of the $i$-th arm at time $t$ $\mat{X}_{t,i} \in \mathbb{R}^{m_1+m_2}$ are the gene expressions of the $i$-th probe for the set of samples for both experimental conditions.
    \vspace{-5pt}
    \item \textit{Reward}: The reward is the observed discrimination provided by the selected gene probe (arm). We define the reward as the logit of the $p$-value of Welch's t-statistic for two-sample t-test to measure discrimination. The reward is defined as 
    \begin{equation}
    \label{eq:probe_selection_reward}
        \hat{r}_t = \log \left ( \frac{1-p_{a_t}}{p_{a_t}} \right) + \epsilon_t = X_{a_t} \beta^* + \epsilon_t
    \end{equation}
    where $p_{a_t}$ denotes the \textit{p-value} of the Welch's test of the null hypothesis that the arm $a_t$ is a non-discriminative gene whose means are identical in each class. 
\end{itemize}

\textbf{Evaluation and Results}: Based on the reward function \eqref{eq:probe_selection_reward}, we apply our contextual bandit and treat the zero expression values in the data $X$ as missing entries. As the goal of the bandit problem is to select the most discriminating DNA probes at each time point, we evaluate the fraction of the probe selections that correctly lead to a statistically significant ($\alpha = 0.05$). To simulate noisy rewards in terms of \textit{p-value}, each probe is sampled with replacement at each time $t$. 

\begin{figure}[ht] \centering
        \includegraphics[width=0.9\linewidth]{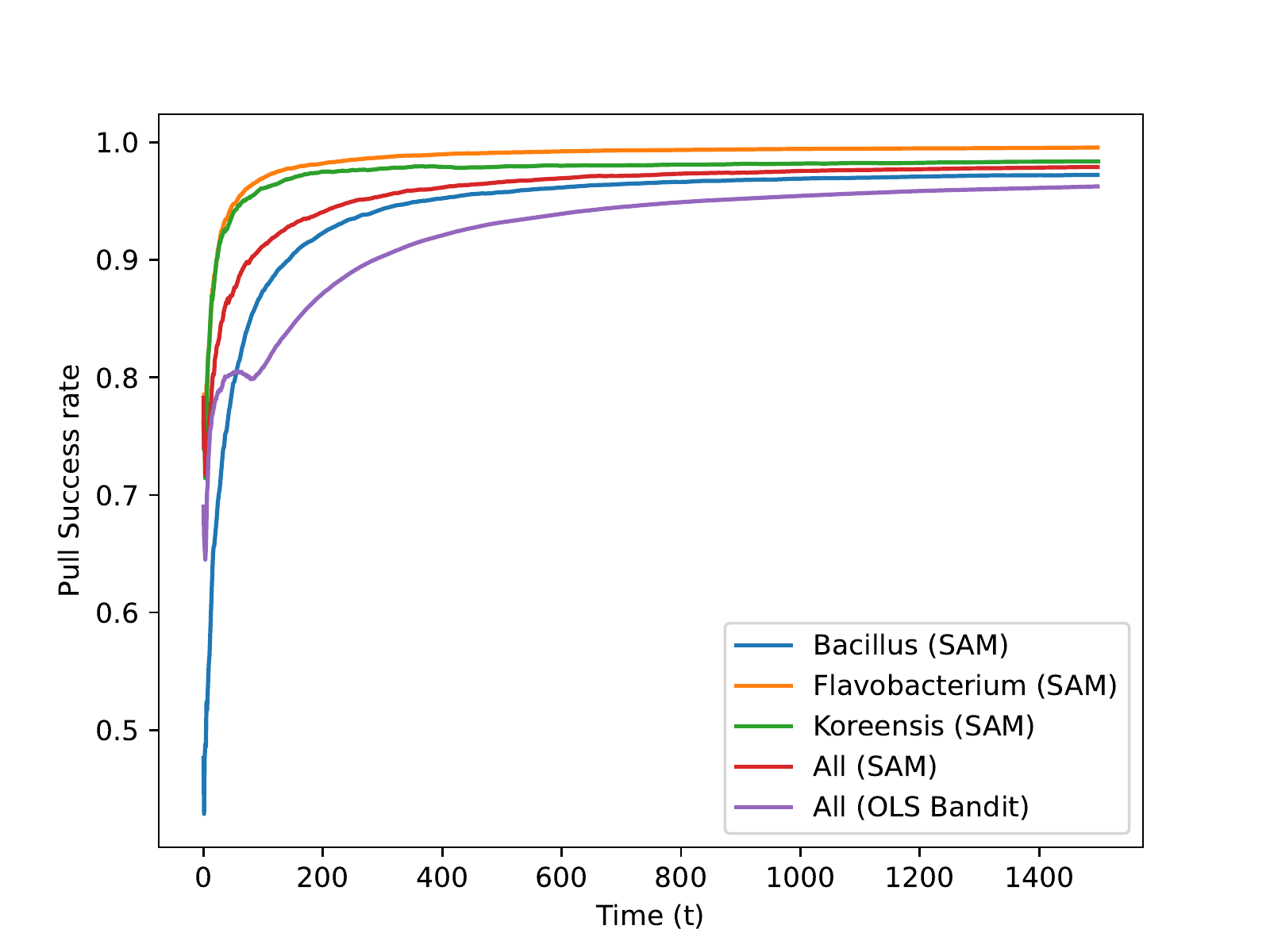}
        \setlength{\belowcaptionskip}{-8pt}
    \caption{Success rate of proposed contextual bandit measured by the fraction of probes selected at each time (arm pull) that highly discriminate between microbiome classes based on \textit{p-value} ($\alpha < 0.05$) of two-sample t-test.
    %Welch's test of significance for testing that class means are identical. 
    Each result represents an average over 100 trials. The proposed sparse agnostic with missingness bandits (SAM) more rapidly achieve 100\% success rate than the standard bandit (OLS).}
    \label{fig:microbial_significance}
\end{figure}

In the early phase, the learner explores the covariate space and selects probes that do not necessarily yield high rewards. After a short period of time, the set of probes selected by the learner quickly narrows down to the subset of statistically significant probes. 
Figure~\ref{fig:microbial_significance} shows the pull success rate for our proposed bandit. For each species, the success rate rapidly reaches around 0.95 for SAM. The difference between the class distributions is the largest for \textit{Flavobacterium} in the overall data, and the bandit quickly discriminates this in its early set of selected probes for $F$ (shown in orange in Figure~\ref{fig:microbial_significance}).

\section{Conclusion}
We introduced a solution to the missing value problem for a sparse linear contextual bandit. Missing covariates often result from the high cost of collecting context vectors or probe sensor failures. A modification of the SA lasso bandit was presented when the context covariates may be incompletely observed. We modeled the problem as missing completely at random but with possibly different covariate missingness probabilities. 
Even in this simple setting, the missingness plug-in lasso estimator results in a non-convex objective function, and we have adopted and extended the regression solution proposed in \cite{lohwainwright12} to bandit problems in this paper.

A natural extensions of our work include 1) the missing-not-at-random framework and 2) the best-k identification problem, which allows multiple arms to be pulled at once. A missingness extension to this problem is still an open problem. 

\bibliographystyle{IEEEbib}
\bibliography{linear_bandit}

\end{document}